# Terrorism Event Classification using Fuzzy Inference Systems

Uraiwan Inyaem
Faculty of Information Technology
King Mongkut's University of Technology North Bangkok
Bangkok, Thailand
uraiwaan@gmail.com

Phayung Meesad
Faculty of Technical Education
King Mongkut's University of Technology North Bangkok
Bangkok, Thailand
pym@kmutnb.ac.th

Choochart Haruechaiyasak
Human Language Technology Laboratory
National Electrics and Computer Technology Center
Pathumthani, Thailand
choochart.haruechaiyasak@nectec.or.th

Dat Tran
Faculty of Information Science and Engineering
University of Canberra
ACT, Australia
dat.tran@canberra.edu.au

*Abstract*—Terrorism has led to many problems in Thai societies, not only property damage but also civilian casualties. Predicting terrorism activities in advance can help prepare and manage risk from sabotage by these activities. This paper proposes a framework focusing on event classification in terrorism domain using fuzzy inference systems (FISs). Each FIS is a decision-making model combining fuzzy logic and approximate reasoning. It is generated in five main parts: the input interface, the fuzzification interface, knowledge base unit, decision making unit and output defuzzification interface. Adaptive neuro-fuzzy inference system (ANFIS) is a FIS model adapted by combining the fuzzy logic and neural network. The ANFIS utilizes automatic identification of fuzzy logic rules and adjustment of membership function (MF). Moreover, neural network can directly learn from data set to construct fuzzy logic rules and MF implemented in various applications. FIS settings are evaluated based on two comparisons. The first evaluation is the comparison between unstructured and structured events using the same FIS setting. The second comparison is the model settings between FIS and ANFIS for classifying structured events. The data set consists of news articles related to terrosim events in three southern provinces of Thailand. The experimental results show that the classification performance of the FIS resulting from structured events achieves satisfactory accuracy and is better than the unstructured events. In addition, the classification of structured events using ANFIS gives higher performance than the events using only FIS in the prediction of terrorism events.

I. INTRODUCTION

The violence of insurgency in three southern provinces of Thailand has increased from 2004 to the present day. The most recognized events were the robbery of rifles at the 4$^{th}$ Development Battalion and the arson of 20 Buddhist schools in Narathiwat province [1, 2]. The insurgents have used terrorism tactics to continue violent attacks which have become daily events. From this problem, many agencies such as official, civilian and expert groups have been given tasks to cautiously handle the events. There is a need for a decision-making solution to support detailed analysis of each terrorism event, for example, to identify which terrorist group conducted the event, which tactic was used for the terrorism event, and who the victims were. It is difficult to affirm or identify the event because each insurgency group and each event are unique and have their own purposes. Moreover, the decision can be binary sense emerging with uncertainty; using technology to support decision-making is an alternative.

Fuzzy inference system (FIS) is a model of decision-making combining fuzzy logic and approximate reasoning. It is generated in five main parts: the input interface, the fuzzification interface, knowledge base unit, decision making unit and output defuzzification interface. FIS is utilized first by Takagi, Sugeno, and Kang [3], and followed Jang [4] applied it to complete this concept for the adaptation with illustrations of various successful applications. Adaptive neuro-fuzzy inference system (ANFIS) is a FIS model adapted by combining the fuzzy logic and neural network in order to determine their properties, fuzzy logic rules and adjustment of MF, by processing dataset samples. ANFIS is a specific approach proposed by Jang [4] which showed significant results in modeling nonlinear functions, and the MF parameters were extracted from a dataset that described the system behavior. The ANFIS learnt features in the dataset and adjusted the system parameters accordingly to a given error criteria. Successful implementations of ANFIS have been reported in many applications [3, 5].

This paper presents a comparison of frameworks of classification using FISs with structured and unstructured events, and a comparison of structured event frameworks of classification using FIS and ANFIS in terrorism domain. The classification runs with terrorism data set of three southern provinces of Thailand, which have been collected from five Thai news publishers from 2007 to 2009. The framework consists of four steps: first, information filtering uses feature selection to filter news articles; second, information extraction uses named entity techniques to extract terrorism events; third, feature formatting transforms text data to the list of vector data; fourth, classification model construction generates model focusing on different FIS setting in unstructured and structured event, and classifies events using FIS and ANFIS, in structured events framework. The correct classification rates of the proposed model are examined and the model performance is reported. The experimental results illustrate FIS resulted from structured events achieves satisfactory accuracy and is better than the unstructured events. In addition, the classification of structured events using ANFIS gives higher performance than the events using FIS in the prediction of terrorism events. The FIS model set from ANFIS obtained from data set achieves satisfactory accuracy which is better than the accuracy of model set from other domain experts on the classification of terrorism events.

The rest of this paper is organized as follows. Section II presents related works for event classification. Section III proposes a framework of event classification in terrorism domain. Section IV describes implementation detail. Section V gives the experimental results and discussion. Final section concludes the paper.

## II. RELATED WORKS

Classification is a process of seeking the correct topic for each document. The main approach to text classification is the knowledge engineering (KE) and machine learning (ML) approach. The KE system usually outperforms the ML system, but the disadvantage of KE is huge amount of highly skilled labor and expert knowledge required to create and maintain knowledge rules. Therefore, most recent works on classification are focused on the ML approach, which require only training set that save cost to produce [6]. Sebastiani [7] surveyed the main approaches to text categorization that within the machine learning paradigm. Fuzzy inference system (FIS) is used to perform text classification by linking fuzzy logic and approximate reasoning. Lee, Jian, andHuang [8] proposed FIS mechanism to construct fuzzy ontology and its application to news summarization. Chao and Teng [9] surveyed an implementing of a FIS using normalized fuzzy neural network. Adaptive neuro-fuzzy inference system (ANFIS) is a FIS model adapted by combining the fuzzy logic and neural network. Recent ANFIS is applied in many applications. Guler and Ubeyli [10] proposed the ANFIS model that combined the neural network adaptive capabilities and the fuzzy logic qualitative approach for detection of electrocardiographic changes in patients with partial epilepsy. Toosi and Kahani [11] surveyed neuro-fuzzy networks, fuzzy inference approach and genetic algorithms for classification system to detect and classify intrusions from normal behaviors based on the attack type in a computer network. Szeifert [12] focused on the design of interpretable fuzzy rule-based classifiers from data with low-human intervention and low-computational complexity. Chakroborty and Kikuchi [13] proposed a way to calibrate the membership function when a set of input and output data was given for the system. Fernandez, Jesus and Herrera [14] proposed work with fuzzy rule based classification systems using adaptive inference system in imbalance data set. Furthermore, this paper contributes a framework for terrorism event classification using fuzzy inference system (FIS) and survey of known FIS techniques and their strengths and weaknesses. However, they could not prove the general validity of their criteria for rule combination to the structure of their fuzzy neural networks. Moreover, no searching algorithm is presented in their paper for searching rules that can be linked. To combine the benefits of fuzzy logic system and a neural network, this paper presents a FIS setting in different ways for event classification terrorism domain to predict a tactic of each terrorism event in three southern provinces of Thailand.

## III. THE PROPOSED FRAMEWORK OF EVENT CLASSIFICATION IN TERRORISM DOMAIN

First of all, news corpus is provided for the research's experiments. After that, the proposed classification framework (in Fig. 1) is described. The framework compoents are information filtering, information extraction, feature formatting, and classification model construction.

### A. News Corpus

The research uses data set which has been collected from five Thai news websites, Thairath, Dailynew, Naewna, Manager and Khaosod from 2007 to 2009. Previous work [15] gathered various news articles on terrorism information from other sources and stored them in a database. The gathered data is terrorism information data set which is used as input for extracting into structured terrorism event information form described as in [16]. The complete data set consists of three types of news articles as follows: 1) Occurrence event article; 2) Finding suspicious item article; and 3) Arrest of suspect article. In this research, the occurrence event articles are the only data set used for training FIS and ANFIS since they can be used to predict tactics of terrorism events.

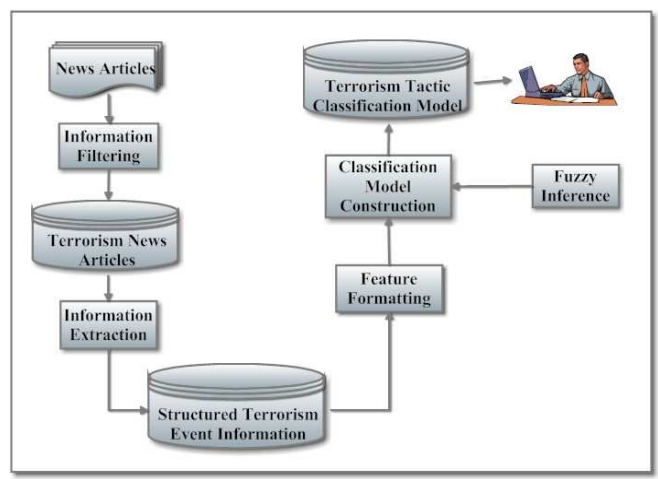

Figure 1. Overview of the proposed framework

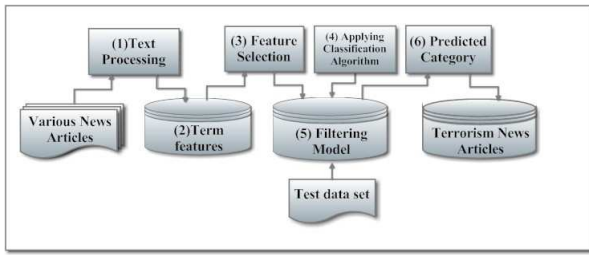

Figure 2.  Terrorism News Article Filtering Architecture

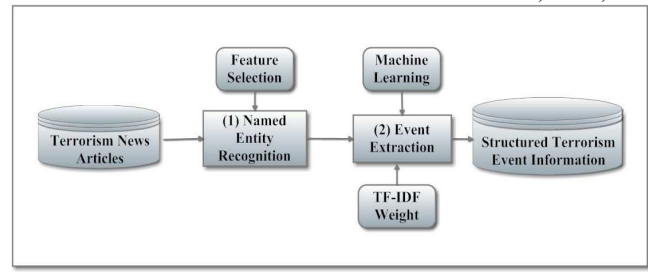

Figure 3. Terrorism News Article Extraction Architecture

Four predefined classes are Demolition tactic, Assassination tactic, Suicide Attack tactic and Unsuccessful Terrorism Event. *Demolition* is to destroy local properties; *Assassination* is to kill officials and populace using ambush tactic; *Suicide Attack* is to destroy local properties, and kill officials or populace, which focuses on the success over terrorist's life, terrorists can sacrifice their life's in the attack [17, 18]; and *Unsuccessful Terrorism Event* means no victim died and damage occurred at lowly important place.

*B. Information Filtering*

The methodology of this research was presented in the previous work [15] which provided a novel system to discriminate terrorism information from other information and stored them in a database. The filtering mechanism is shown in Fig. 2. The information filtering methodology is as follows. First, the initial process starts with text processing module, i.e. (1), which the Thai news articles are manually classified into sets of category labels. The segmentation algorithm transforming texts to word tokens is the major task of text processing for Thai text. The word segmentation program called LexTo[1], which is the basic longest matching open source, is used in this experiment. Next process, two sets of extracted words, i.e., (2) that are unknown-word set and known-word set are obtained from the news corpus. The experiment focuses in terrorism domain, which compares the performance of the word segmentation result between Thai electronic dictionary, LEXiTRON[2] and Domain Knowledge (DKG) dictionary. To create DKG dictionary, human Thai language expert analyzes the meaning of the word from unknown-word set, this procedure happens repeatedly until all unknown words are removed from the segmentation processing a result set. The resulting set of extracted words is stored as term words in the database.  After that, a feature selection process, i.e., (3), is applied to the feature selection technique as in [15], such as information gain (IG), chi squared (CHI), relief, and correlation based feature selection (CFS), they reduce the number of word features remaining only necessary words. The result is a set of feature vectors that is ready to be used by the text categorization algorithms. Each feature is a unique word which appears in the training collection. This paper used Normalized Correlation based Feature Selection (NCFS) as the weighting scheme for assigning the feature values. Later, applying classification algorithm, i.e., (4), is to apply classification to construct classification model. Classification algorithms [15], linear regression (LR), decision tree (DTREE), naïve bayes (NB), and support vector machine (SVM) are applied to build models by using the feature vectors. Next process, (5) filtering model, this filtering model is used to account the usefulness of individual feature scores for each category based on the test feature vector. Final process, test data is first transformed into a feature vector and is predicted into a predefined category by using filtering model.. The test data is assigned to the category whose score is the maximum among them. The resulting news articles are stored in the database called terrorism news articles.

*C. Information Extraction*

The result of the previous process is the terrorism news articles corpus used for input in this process, the Thai terrorism event extraction as in [16, 19]. The system can be viewed as a pipeline of process shown in Fig. 3. The initial process, the corpus is preprocessed before processing of the named entity recognition.  Next, a terrorism information event extraction process is performed by comparing four machine learning algorithms as in [16], namely SVM, NB, k-nearest neighbor (KNN), and DTREE. Finally, the extracted events are stored into element template pattern. Each step is discussed in more details below.

*1) Named-Entity Recognition*

A pre-process of the corpus is executed to obtain named entity feature selection for learning. The corpus of the Thai terrorism news article is prepared by exporting it into a single text file as Extensible Markup Language (XML) format [20]. These training files have to annotate some news articles with labels that are about 70% of the total article, which want the learning system to annotate. The remaining, 30% of the articles are included as test files. News article header and body are demarcated with XML tags, <header> and <content>. The number of annotated entities in this task includes up to nineteen types which are Day, Month, Year, Time, Period, Suburb, District, City, Tactic, Evidence, Terrorist's Name, Status of Terrorist, Level of Terrorist, Place, Title, Victim's Name, Status of Victim, Victim's Occupation, and Amount of Weapons.  The linguistic feature selections of this task are compared and studied, such as terrorism gazetteer, terrorism ontology and terrorism grammar rule according to the rules-based recognizer [16].

---

[1]  LexTo: Thai lexeme tokenizer, *http://www.sansarn.com*
2 LEXiTRON dictionary, *http://lexitron.nectec.or.th*

*2) Event Extraction*

Machine learning is used to extract terrorism events from the Thai terrorism news article with linguistic feature selection using the GATE tool. The tool relies on finite state algorithms and the JAPE language, as in [21]. The task mainly involves under taking three subtasks. The first is to annotate some news articles with the class labels that it wants the learning system to annotate in some news articles. The second is to prepare the corpus to obtain the linguistic feature selection for the learning system. Finally, an XML configuration file for setting the machine learning application interface is created. Algorithms are implemented in WEKA[3] (the Waikato Environment for Knowledge Analysis). Each term feature is represented by the term frequency-inverse document frequency (TF-IDF). Finite State Transduction is applied for learning feature weights. A training set builds a model, and a test set to run on the new documents. The results are extracted and saved to database.

*D. Feature Formatting*

Structured terrorism events are the resultant data from the information extraction. This research used terrorism events for the input in ANFIS. The input format is the list of vector data values, so the text data formatting is changed to an integer formatting. The feature formatting is presented to transform the structured terrorism events into an integer formatting, and store them with the list of vector as shown in Fig. 4. Feature formatting procedure:

1. *pull* the requirement of three element from the structured terrorism event storage such as place, the status of victim, the status of terrorist
2. *define* the priority of three parameters
3. *define* the meaning for text data
4. *replace* the text number with an integer using function script
5. *recheck* over all again

The result, *[1 1 1 1 ]*, is a list of vector data that consists of 4 vector columns. Each column meaning is place, the status of victim, the status of terrorist, and class label, respectively as shows in Fig. 5.

*E. Classification Model Construction*

A proposed framework has main four steps: 1) information filtering; 2) information extraction; 3) feature formatting; and 4) classification model construction.

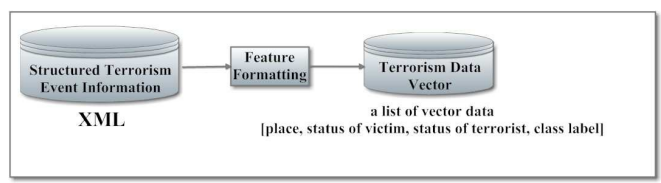

Figure 4. The feature formatting process

---
[3] WEKA, *http://www.cs.waikato.ac.nz/ml/weka.*

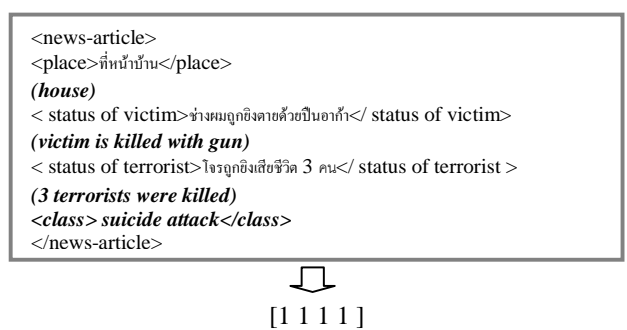

Figure 5. An example of the feature formatting

This paper proposed fuzzy inference systems (FISs) for classification model construction. The construction used FIS to generate models in different way. Three approaches are presented as in Fig. 6. Each approach detail is described as follows:

- *Approach 1 (Unstructured)* : Unstructured terrorism events are an input for FIS to generate classification model. The step of this approach as follows: Terrorism news articles or unstructured terrorism events, information filtering, feature formatting, defined fuzzy inference, classification model construction, classification model, testing data, performance evaluation, and error report.
- *Approach 2 (Structured)* : The process of this approach is similar to *Approach 1*, but this model uses structured terrorism event as an input for FIS to construct classification model. Structured terrorism events are derived from an extraction process.
- *Approach 3 (ANFIS)* : This approach is extended from *Approach 2* by using ANFIS to build classification model.

The comparisons of the proposed approaches in two ways: first is the comparison between unstructured events (approach 1) and structure events (approach 2) using the same FIS setting; second comparison is the model settings comparing between structured events classified by FIS (approach 2) and ANFIS (approach 3).

Fuzzy interface system (FIS) [22] under this consideration has three inputs $x_1$, $x_2$ and $x_3$ and each input has three linguistic terms, for example, {$A_1$, $A_2$, $A_3$} or {*low, medium, high*} for input $x_1$, so there are 27 fuzzy if-then rules as following :

**Rule 1** : If $x_1$ is $A_1$ and $x_2$ is $A_2$ and $x_3$ is $C_1$, then $f_1 = p_1^1 x_1 + p_2^1 x_2 + p_3^1 x_3 + p_4^1$

**Rule 2** : If $x_1$ is $A_2$ and $x_2$ is $B_2$ and $x_3$ is $C_2$, then $f_2 = p_1^2 x_1 + p_2^2 x_2 + p_3^2 x_3 + p_4^2$

…..

**Rule 27** : If $x_1$ is $A_{27}$ and $x_2$ is $B_{27}$ and $x_3$ is $C_{27}$, then $f_{27} = p_1^{27} x_1 + p_2^{27} x_2 + p_3^{27} x_3 + p_4^{27}$

where superscript denotes the rule number. Fig. 7. shows five layer network of FIS structure. A brief explanation for construction of the classification model is described. For first order, three inputs fuzzy model (j) is a common fuzzy if-then rule. The output of the $i^{th}$ node in layer $k$ is $O_{k,i}$. Every node in *Layer 1* can be any parameterized MF, such as the generalized triangular MF. These parameters are called *premise parameter* which are nonlinear coefficients. The fixed node function in *Layer 2* is a output the product of all inputs, and the output stands for the firing strength of the corresponding rule.

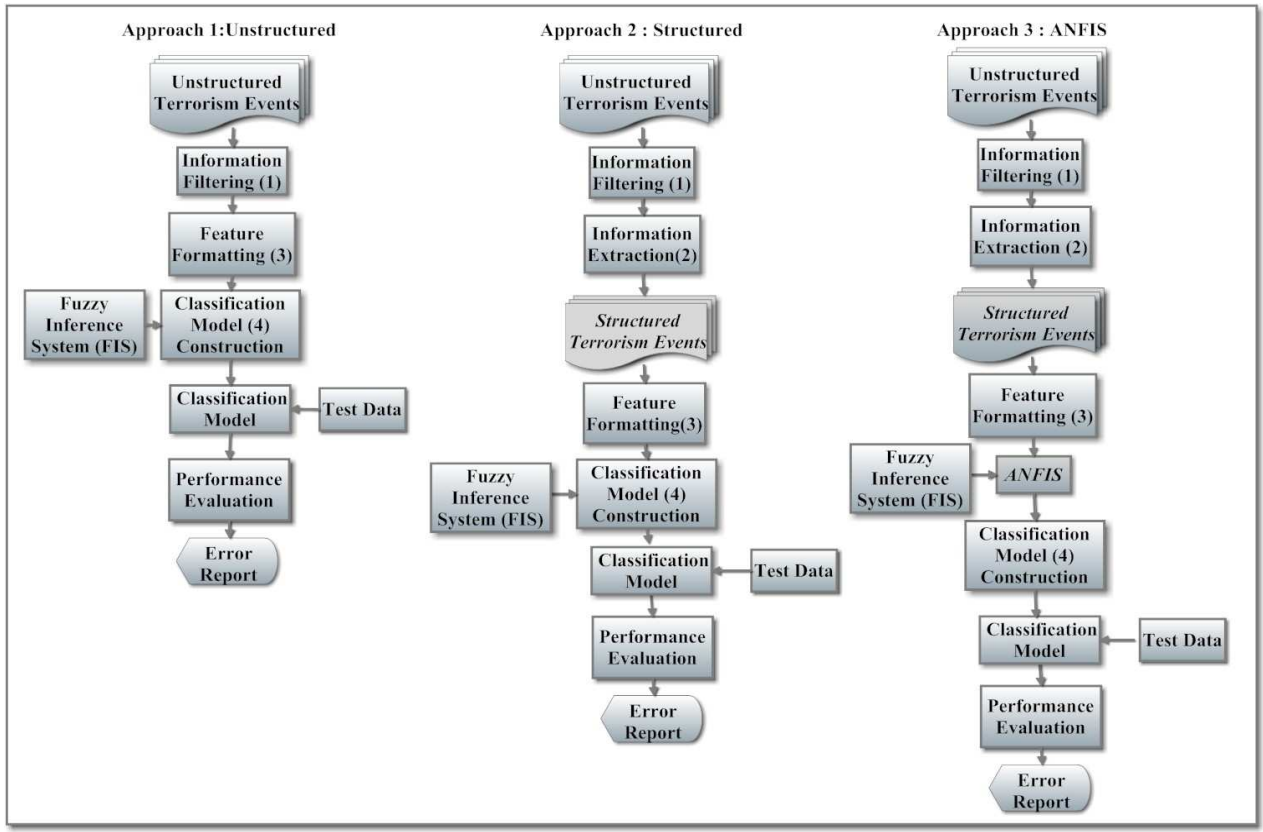

Figure 6. Three proposed approaches

$$O_{2,i} = w_i = \mu_{A_j}(x_1)\mu_{B_j}(x_2)\mu_{C_j}(x_3), \quad i=1,2,...,27; \quad j=1,2,3 \quad (1)$$

The fixed node function in *Layer 3* is used to normalize the input firing strengths.

$$O_{3,i} = \bar{w}_i = \frac{w_i}{\sum_{i=1}^{27} w_i}, \quad i=1,2,...,27 \quad (2)$$

Every node in *Layer 4* is a parameterized function, and the adaptive parameters are called *consequent parameters*. The node function is given by:

$$O_{4,i} = \bar{w}_i f_i = \bar{w}_i\left(p_1^i x_1 + p_2^i x_2 + p_3^i x_3 + p_4^i\right), \quad i=1,2,...,27 \quad (3)$$

There is only one node in *Layer 5* with a simple summing function.

$$O_{5,i} = \sum_i \bar{w}_i f_i \quad (4)$$

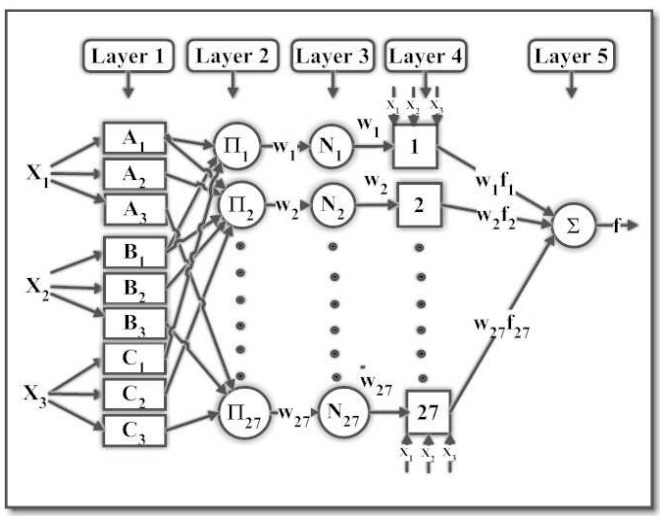

Figure 7. The FIS processing structure

Therefore, the ANFIS network is constructed corresponding to FIS model. The ANFIS architecture can be update its parameters according to the neural network algorithm. The hybrid learning algorithm in ANFIS, considering as in (2) and (3), the values of premise parameter of the overall output are a linear combination of the consequence parameters. The learning has been divided into forward and backward passes [22, 23, 24]. Table I. described the forward pass of the learning algorithm stop at nodes in *Layer 4* , and the consequent parameters are identified by the least squares method. For the backward pass, the error signals propagate backward and the premise parameters are undated by gradient descent. The MF is fixed and only the consequent part is adjusted [4], so the ANFIS can be viewed as a functional-link network, where the enhancement node functions of the input variables is achieved using the MF. The only difference here is that the enhancement function in this fixed ANFIS which takes advantage of the human knowledge, which is more revealing of the problem, than function expansion or random generated function.

Table I. The learning algorithm for ANFIS

| Part | Forward Pass | Backward Pass |
|---|---|---|
| **Premise parameters** | Fixed | Gradient descent |
| **Consequent parameters** | Least-squares | Fixed |
| **Signals** | Node outputs | Error signals |

## IV. IMPLEMENTATION

Before the experiments are set up, we must define components of fuzzy inference system (FIS) for classification model construction. Previous section describes the principal of FIS work, and then the proposed FIS is presented and implemented in MATLAB[4] program. FIS produces in five layers as follows: 1) input interface; 2) fuzzification interface; 3) knowledge base unit; 4) decision making unit; and 5) output defuzzification interface are described below.

*Layer 1 Input Interface:* performs input processes. The element of events is transformed into a list of vector data provided by feature formatting to directly transmit input value to next layer. The input vectors are the element set retrieved from the element transformation as shown in Fig. 5.

*Layer 2 Fuzzification interface:* the membership function (MF) is computed the membership degree for all terms. The selection of MF is very impacted on the system performance, and usually depends on their specific problem [25]. This is often determined heuristically and subjectively. In this paper, three approaches are investigated with MF and their parameters [26, 27]. Eight types of MF are used for fuzzy system modeling. They are Triangular, Trapezoidal, Gaussian, Two-Side Gaussian, Bell-Shape, Product of Two Sigmoid, Differences Between Two Sigmoid and Pi-Shape, respectively [20]. They are defined as follows:

*1) Triangular MF:* The expression is shown in (5).

$$\Delta(x;a,b,c) = \begin{cases} 0 & \text{when } x \leq a, x \geq c \\ (x-a)/(b-a) & \text{when } a \leq x \leq b \\ (c-x)/(c-b) & \text{when } b \leq x \leq c \\ 0 & \text{when } c \leq x \leq d \end{cases} \quad (5)$$

where the parameters *a, b,* and *c* determine the shape triangular MF.

---
[4]MATLAB, available at http://www.mathworks.com/

*2) Trapezoidal MF:* the expression is shown in (6).

$$\Pi(x;a,b,c,d) = \begin{cases} 0 & \text{when } x \leq a, x \geq d \\ (x-a)/(b-a) & \text{when } a \leq x \leq b \\ 1 & \text{when } b \leq x \leq c \\ (d-x)/(d-c) & \text{when } c \leq x \leq d \end{cases} \quad (6)$$

where the parameters *a, b, c* and *d* control the shape of trapezoidal MF.

*3) Gaussian MF:* the expression is defined in (7).

$$G(x;m,\sigma) = e^{-(x-m)^2/2\sigma^2} \quad (7)$$

where the parameters $m$ and $\sigma$ determine center and width of the MF.

*4) Two-Side Gaussian MF:* the expression is defined in (8).

$$\psi(x;m_1,m_2,\sigma_1,\sigma_2) = \begin{cases} e^{-(x-m_1)^2/2\sigma_1^2} & \text{when } x \leq m_1 \\ e^{-(x-m_2)^2/2\sigma_2^2} & \text{when } x \geq m_2 \end{cases} \quad (8)$$

where the parameters $m_1, \sigma_1$ and $m_2, \sigma_2$ determine the shape of two-side Gaussian MF and correspond to the centers and widths of the left and right half Gaussian functions.

*5) Bell-Shape MF:* This function depends on three parameters namely *a, b,* and *c* as given by (9).

$$\Omega(x;a,b,c) = e^{-1/1+|(x-c)/a|^{2b}} \quad (9)$$

where the parameter *a* and *b* are usually positive and the parameter *c* locates the center of the curve.

*6) Product of Two Sigmoid MF:* This function depends on four parameters namely $m_1, m_2, c_1, c_2$ as given in (10).

$$\tau(x;m_1,c_1,m_2,c_2) = 1/\left(1+e^{-m_1(x-c_1)}\right)\left(1+e^{-m_2(x-c_2)}\right) \quad (10)$$

where the parameters $m_1, c_1$ and $m_2, c_2$ determine the shape of two sigmoid MF.

*7) Difference Between Two Sigmoid MF:* The function depends on four parameters namely $a_1, a_2, m_1, m_2$ given in (11).

$$\beta(x;a_1,m_1,a_2,m_2) = \left\|\left[1/(1+e^{-a_1(x-m_1)})\right]-\left[1/(1+e^{-a_2(x-m_2)})\right]\right\| \quad (11)$$

where the parameters $a_1, m_1$ and $a_2, m_2$ determine the shape of difference between two sigmoid MF.

*8) Pi-Shape MF:* This MF is a product of Z shape and S shape function. The MF is evaluated at the points determined by the vector x. The parameter *a* and *d* represent the feet of the curve, while *b* and *c* are its shoulders, which depend on four parameters namely *a, b, c* and *d* as given in (12).

$$\pi(x; a, b, c, d) = \begin{cases} 0, & x \leq a \text{..and..} x \geq d \\ 2\left(\frac{x-a}{b-a}\right)^2, & a \leq x \leq \frac{a+b}{2} \\ 1-2\left(\frac{x-b}{b-a}\right)^2, & \frac{a+b}{2} \leq x \leq b \\ 1, & b \leq x \leq c \\ 1-2\left(\frac{x-c}{d-c}\right)^2, & c \leq x \leq \frac{c+d}{2} \\ 2\left(\frac{x-d}{d-c}\right)^2, & \frac{c+d}{2} \leq x \leq d \end{cases} \quad (12)$$

In (5) to (12) shows the MFs of the fuzzy sets for all variables in this research. The value of membership degree is usually in the range [0, 1] where "1" means full membership and "0" means no membership. For fuzzy variables, there are *place*, *the status of victim* and *the status of terrorist*. These inputs are considered for each of their properties, and are performed the membership degree computing. Table II shows the structure of three fuzzy variables namely *place*, *the status of victim* and *the status of terrorist* fuzzy variable with three linguistic terms, such as *low*, *medium* and *high.* Table III shows the *tactic* class labels that have four linguistic terms namely *very low, low*, *medium* and *high,* respectively. Table IV domain experts define the minimum, mean and maximum of bound for all fuzzy variables and fuzzy values set as follow: 1) *place* variable; 2) *the status of victim* variable; 3) *the status of terrorist* variable; and 4) *Tactic* variable (class labels). First three fuzzy variables have fuzzy value sets namely *low*, *medium* and *high*, and last fuzzy variable has also *very low* value set. All fuzzy values of MF are not the same because they are computed using the different MFs and different parameters. The fuzzy value depends on definition of the bound fuzzy variable that is defined by domain experts.

*Layer 3 Knowledge Base Unit:* the fuzzy rules are represented. The basic idea is to perform precondition matching of fuzzy logic rules with their links, so the outputs are linked with associated linguistic term in the next layer. The algebraic product operation is used to compute the precondition matching degree as shown in (13).

Table II. Fuzzy values of fuzzy variables defined by domain expert

| Fuzzy Values Stand | Fuzzy Values Abbreviate | Place Meaning | The status of victim | The status of terrorist |
|---|---|---|---|---|
| low | L | Lowly important place | No victim died | No terrorist died |
| medium | M | Important place | Victim injured | Terrorist injured |
| high | H | Very important place | Victim died | Terrorist died |

Table III. Class labels of *tactics* defined by domain expert

| Class label Stand | Class label Abbreviate | Tactic Stand | Tactic Meaning |
|---|---|---|---|
| very low | VL | Unsuccessful attack or no important damage | No victim died and damage occurred at lowly important place |
| low | L | Demolition | local properties destroyed |
| medium | M | Assassination | Officials or populace killed |
| high | H | Suicide attack | local properties, destroyed officials or populace killed, focusing on the success over terrorists' life |

Table IV. The fuzzy variables, values, and their range are defined by domain experts.

| No. | Fuzzy Variable | Fuzzy Values | Left Bound | Mean Bound | Right Bound |
|---|---|---|---|---|---|
| 1 | Place (P) | P_low | 0 | 1.0 | 9.0 |
|  | Range [ 0-25] | P_medium | 6.0 | 13 | 20.0 |
|  |  | P_high | 16 | 25 | 25 |
| 2 | The status of victim (V) | V_low | 0 | 0 | 1.2 |
|  | Range [ 1-4] | V_medium | 0.3 | 1.5 | 2.7 |
|  |  | V_high | 1.8 | 3 | 4.0 |
| 3 | The status of terrorist (T) | T_low | 0 | 0 | 1.2 |
|  | Range [ 1-4] | T_medium | 0.3 | 1.5 | 2.7 |
|  |  | T_high | 1.8 | 3 | 4.0 |
| 4 | Tactic (TT) | Tt_verylow | 1 | 1 | 2.2 |
|  | (class label) | Tt_low | 1.3 | 2.5 | 3.7 |
|  | Range [ 1-4] | Tt_medium | 1 | 2.5 | 4 |
|  |  | Tt_high | 2.8 | 4 | 4 |

In this paper, domain experts define the rules as shown in Table V.

$$O_{ij} = \min\left(t_{1i\_place\_low}, t_{1i\_status\_of\_victim\_low}, t_{1i\_status\_of\_terrorist\_low}\right) \quad (13)$$

*Layer 4 Decision-Making Unit:* the multiplier receives the incoming signals and sends the product out as a circle node. Each node output represents the firing strength of a rule. T-norm operators perform in this layer (AND, OR, NOT). The fuzzy values of outputs are defined by domain experts, which is four linguistic terms namely *very low*, *low*, *medium*, and *high* as shows in Table V. respectively.

*Layer5 Output Defuzzification interface:* performing the defuzzification process to get the *tactic* of the term pair. The center of area method [8] is adopted to carry out the defuzzified process. The term pair of event is computed for the *tactic* values using (14) as below.

$$O_{ij} = \frac{\sum_{p=1}^{r}\sum_{q=1}^{c} O_{pq} V_{pq}}{\sum_{p=1}^{r}\sum_{q=1}^{c} O_{pq}} \quad (14)$$

where *r* is the number of rule nodes, *c* is the number of linguistic terms of output fuzzy variable, and $V_{pq}$ is the gravity of $q^{th}$ output linguistic term associated with the $p^{th}$ rule node. This research defined *r* and *c* to be 27 and 4, respectively.
The summation of output class is computed using Eq. (15).

$$OC_j E_1 = \frac{\sum_{i=1}^{n_1} O_{ij}}{n_1} \quad (15)$$

For all *j*=1,…,*m*

Next is to perform integration the membership degree belonging to all news articles represented in this form below.

$$\{(C_1; O_{C_1 E_1}, \mu_{C_1 E_1}, \ldots, \mu_{C_1 E_p})$$
$$\ldots$$
$$(C_m; O_{C_m E_1}, \mu_{C_m E_1}, \ldots, \mu_{C_m E_p})\} \quad (16)$$

Table V. Fuzzy inference rules defined by domain experts

| Rules | Fuzzy variables | | | Class labels |
|---|---|---|---|---|
| | Place (P) | the status of victim (V) | the status of terrorist (T) | Tactic (TT) |
| 1 | H | H | H | H |
| 2 | H | H | M | M |
| 3 | H | H | L | M |
| 4 | H | M | H | H |
| 5 | H | M | M | L |
| 6 | H | M | L | L |
| 7 | H | L | H | H |
| 8 | H | L | M | L |
| 9 | H | L | L | L |
| 10 | M | H | H | H |
| 11 | M | H | M | M |
| 12 | M | H | L | M |
| 13 | M | M | H | H |
| 14 | M | M | M | M |
| 15 | M | M | L | M |
| 16 | M | L | H | H |
| 17 | M | L | M | L |
| 18 | M | L | L | L |
| 19 | L | H | H | H |
| 20 | L | H | M | M |
| 21 | L | H | L | M |
| 22 | L | M | H | H |
| 23 | L | M | M | M |
| 24 | L | M | L | M |
| 25 | L | L | H | VL |
| 26 | L | L | M | VL |
| 27 | L | L | L | VL |

Moreover, adaptive neuro-fuzzy inference system (ANFIS) is an extension of the TSK fuzzy model. In this paper, ANFIS toolbox is available in MATLAB program. ANFIS used neural network to identify parameters. Neural network in this research adapts back-propagation (BP) learning algorithm and hybrid algorithm that combine the gradient method and the least square method estimation (LSM) for identification parameters. It adjusts the shape parameters of membership function (MF) in FIS by learning from a set of given input and output data. In general ANFIS is more complicate than FIS; the design of itself considers the features specification as follow:

*1) Learning and reasoning*: ANFIS algorithm provides a leaning method of neural network (NN) which can obtain corresponding fuzzy rules from data for fuzzy modeling. The learning method can effectively calculate the best parameter for the MF and best simulate the expected or actual relations between inputs and outputs based on given data.

*2) Structure and parameters*: The ANFIS structure similar to the structure of neural network (NN) which the input space are mapped into output space with their MFs and parameters. For the parameters, the ANFIS function estimates MFs using NN, namely back-propagation (BP) or hybrid algorithm that combine the least square method (LSM) estimation with the gradient method. The parameters can adjust and change the shape of MFs to reduce error between the desired output (target) and the expected system (FIS output) through a learning procedure. *Parameter* in each MF is investigated, such as the training epochs, the number of MFs for each input, the optimization method, the type of output MFs, and the training data and testing data.

*3) Model and data set*: The modeling procedure starts to connect input variables, the MFs of input variables, fuzzy rules, output variables and the MFs of output variables. Next, the results contained a group of pairs of input and output data forms training data to the algorithm of ANFIS according to a certain format. Finally, the process generates the model continuously simulating it to the given training data.

*4) Measure performance:* The measurement of the result model measures training data simultaneously. The function chooses suitable results according to the rules resulting in the least error in the training data, which calculates and compares correspondingly. If the system meets a case of no match, the error of the training data becomes less, so that means the system parameters will not match with the training data.

V. RESULTS AND DISCUSSION

*A. Performance Evaluation*

The performance measures use root mean squared error (RMSE) which is the standard deviation of regression, for evaluation the experiment of terrorism information classification as shown in (17) and (18).

$$RMSE = \sqrt{MSE} \quad (17)$$

$$MSE_{(i)} = \frac{(N-p)MSE - e_i / (1 - H_{ii})}{N - p - 1} \quad (18)$$

These equations are calculated from the data set, where *i* is the estimated parameter, N is data size, respectively. A value of the root mean squared error closes to 0 indicates a better fit.

*B. Experimental results*

The experiment compares FIS settings into two comparisons: first is the comparison between unstructured events (approach 1) and structured events (approach 2) using the same FIS setting; second comparison is the model settings comparing between structured events classified by FIS (approach 2) and ANFIS (approach 3). The eight MFs, such as Triangular, Trapezoidal, Gaussian, Two-Side Gaussian, Bell-Shape, Product of Two Sigmoid, Differences Between Two Sigmoid, and Pi-Shape are tested using training set option. The results in terms of RMSE measure are averaged across all training set experiments. The experiments compare the performance of the different MF. The details of experiment results are summarized in Table VI. The first comparison result shows the classification of FIS resulted from structured events achieves satisfactory accuracy, the classification error of structured event 2.16 %, and is better than the unstructured events, the classification error of unstructured event 3.92 %, in Bell-Shape MF. The second comparison result shows the classification of structured events using ANFIS gives higher performance, the classification error 0.08 %, than the structured events using FIS, the classification error 2.16 %, in Trapezoidal MF.

Table VI. Experimental results.

| Membership Function Name | Classification Error (RMSE %) | | | | | |
|---|---|---|---|---|---|---|
| | Approach 1: Unstructured | | Approach 2: Structured | | Approach 3: ANFIS | |
| | Train | Test | Train | Test | Train | Test |
| Triangular | 4.54 | 4.53 | 2.42 | 2.42 | 0.09 | 0.09 |
| Trapezoidal | 4.43 | 4.43 | 2.32 | 2.22 | **0.08** | **0.08** |
| Bell-Shape | **3.93** | **3.92** | **2.16** | **2.16** | 0.09 | 0.09 |
| Gaussian | 4.54 | 4.54 | 2.50 | 2.20 | 0.10 | 0.10 |
| 2 Side Gaussian | 4.43 | 4.43 | 3.11 | 3.81 | 0.10 | 0.10 |
| Pi-Shape | 4.43 | 4.42 | 3.22 | 3.22 | 0.14 | 0.14 |
| Different 2 Sigmoid | 4.43 | 4.43 | 2.81 | 2.81 | 0.10 | 0.10 |
| Product of 2 Sigmoid | 4.43 | 4.43 | 2.81 | 2.81 | 0.10 | 0.10 |

*C. Discussion*

This part discusses to clear the understanding of types of terrorism news in this research's experiments. The research separates terrorism news into two groups which are approved news and non-approved news. Approved new must be affirmed by reliable sources while non-approved news is general news without any conformation or is confirmed by a non-reliable source. Fig. 8, after approved news is processed in predicting process, it will be added to the reliable database. It means this news can be believed and can be used to be a database for predicting terrorist's behavior to conduct decision-making to prevent violent events. For non-approved news, the extracted news from this type of news is input to the predicting process, terrorist's tactics can only be added to the unreliable database. They need to be confirmed by domain experts in order to change the status from unreliable to reliable data.

VI. CONCLUSION

This paper compares the performance of classification framework based on FISs for structured and unstructured terrorism events. In addition, we compared between two classification frameworks based on FIS and ANFIS. The data set is a collection of news articles related to terrorism events in three southern provinces of Thailand. The experimental results showed that the classification of FIS using the bell-shape MF yielded the classification error (RMSE) of 2.16 % for structured events, and 3.92 % for unstructured events. In addition, the classification of structured events using ANFIS gives higher performance, with the classification error of 0.08 %, than FIS which gives the classification error of 2.16 %, based on the trapezoidal MF. In conclusion, using ANFIS is the best alternative for event classification to support decision-making for terrorism event prediction.

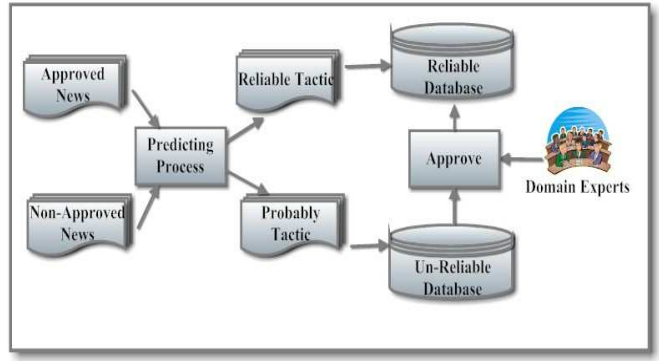

Figure 8. Terrorism event prediction process

## AUTHORS PROFILE

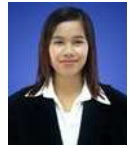

**Uraiwan Inyaem** received the B.Sc. degree in Computer Science from the Rajamangala University of Technology Thunyaburi, Pathumthani, Thailand in 1997, the M.Sc. degree in Information Science from the King Mongkut's Institute of Technology Ladkrabang, Bangkok, Thailand in 2002. She is currently a Ph.D. student candidate in the department of Information Technology at King Mongkut's University of Technology North Bangkok, Thailand. She had done her research in the Faculty of Information Sciences and Engineering at University of Canberra, Australia during January to December 2009. Her current research interests text mining, fuzzy ontology application and natural language processing.

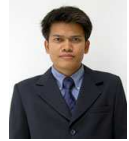

**Phayung Meesad** received the B.Sc. from King Mongkut's University of Technology North Bangkok, Thailand, the M.Sc. and Ph.D. degree in Electrical Engineering from Oklahoma State University, United States of America. Currently, he is an Assistant Professor in Department of Teacher Training in Electrical Engineering at King Mongkut's University of Technology North Bangkok, Thailand. His current research interests include fuzzy systems and neural networks, evolutionary computation and discrete control systems.

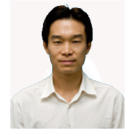

**Choochart Haruechaiyasak** received B.Sc. from University of Rochester, M.Sc. from University of Southern California and Ph.D. degree in Computer Engineering from University of Miami, United States of America. Currently, he is chief of the Intelligent Information Infrastructure Section under the Human Language Technology Laboratory (HLT) at National Electronics and Computer Technology Center (NECTEC), Thailand. His current research interests include search technology, data/text/web mining, information filtering and recommender system.

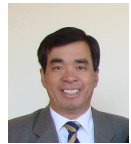

**Dat Tran** received the B.S. and M.S. degrees in Theoretical Physics from University of Science in 1984 and 1994, respectively. He then received Graduate Diploma and Ph.D. degree in Information Sciences and Engineering from University of Canberra, Australia in 1996 and 2001, respectively. He was awarded an internship to work at IBM Watson Research Center, New York, from June 2000 to Sep 2000. Currently, he is Associate Professor at University of Canberra, Faculty of Information Sciences and Engineering, Australia. His research interests include machine learning, data mining, pattern recognition, biometrics authentication, network security, spam email filtering, and fuzzy set theory.